\documentclass[letterpaper, 10 pt, conference]{ieeeconf}  

\usepackage{amssymb}
\usepackage{booktabs}
\usepackage{graphicx} 
\usepackage{xspace}
\usepackage{cite}
\usepackage{amsmath,amssymb,amsfonts}
\usepackage{textcomp}
\usepackage{xcolor}
\usepackage{balance}
\usepackage{algpseudocode}
\usepackage{caption}
\usepackage{float}
\newcommand{\system}{STEP\xspace}

\usepackage{subcaption}
\usepackage{placeins}

\IEEEoverridecommandlockouts                              

\title{STEP: State-Aware Task Estimation and Planning with Multi-Modal LLMs for Human-Robot Collaboration}

\author{}
\author{Maitrey Gramopadhye$^{1}$, Prakash Baskaran$^{2}$, Xiao Liu$^{2}$, Songpo Li$^{2}$ and Soshi Iba$^{2}$
\thanks{This work was done during an internship at Honda Research Institute}
\thanks{$^{1}$University of North Carolina at Chapel Hill, United States}%
\thanks{$^{2}$Honda Research Institute, San Jose, CA, USA}%
\thanks{Correspondence email: maitrey@cs.unc.edu}%
}

\begin{document}

\maketitle
\thispagestyle{empty}
\pagestyle{empty}

\begin{figure*}[ht]
  \centering
  \includegraphics[width=0.85\linewidth]{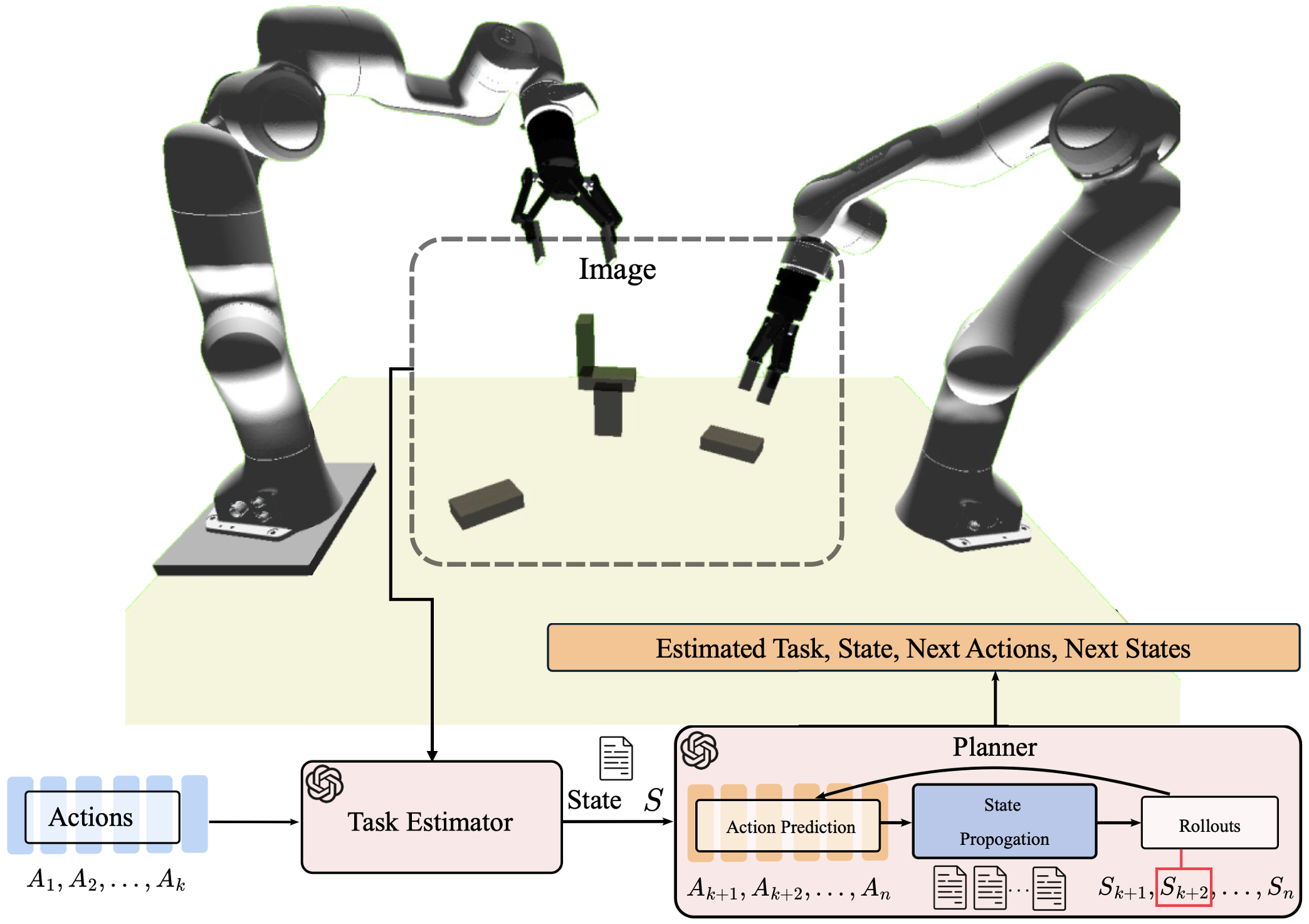}
  \caption{\textbf{\system with robot system setup.} Predicting structured state representations along with future actions allows \system to quantitatively measure the distance of a state from the goal and deduce assistance parameters required for executing the predicted actions.}
  \label{fig:teaser}
  \vspace{-0.2cm}
\end{figure*}

\begin{abstract}
\label{sec:abstract}
Effective human-robot collaboration in industrial settings requires robots to understand human intentions and assist with task planning, reducing workload. 
Recent works have explored the use of Multi-modal Large Language Models (MM-LLMs) for task planning in such data-scarce scenarios, leveraging in-context learning to interpret user actions and generate long-horizon action plans in natural language. 
However, MM-LLMs inherently lack an understanding of system states and do not track state transitions, often leading to hallucinated actions that deviate from the intended goal. Additionally, generating action plans in natural language tends to limit the generated plans to a high level, introducing ambiguity in action execution.
To address these limitations, we propose the State-aware Task Estimator and Planner (\system), which prompts a MM-LLM to explicitly estimate the state of the system and predict the state transitions resulting from executed actions. By forecasting future states alongside actions, \system ensures task-convergent planning while also providing additional assistance parameters necessary for executing the predicted actions.
We evaluate \system in a simulated environment using a robot assembly task. Our approach outperforms the state-of-the-art by $32.8\%$ in action executability and $14.8\%$ in final-state error.










\end{abstract}



\section{Introduction}
\label{sec:intro}
As robots have become more affordable and versatile, their applications have expanded, particularly in industries requiring greater human interaction and generalization.
Human-robot collaboration improves efficiency by drawing on their respective strengths. 
However, effective teamwork requires shared control and mutual understanding. 
For seamless collaboration, robots must be able to infer human intentions and take over some of the workload. 
Relieving control over to a robot can enable a person to supervise multiple robots or take on a creative role by developing a vision and delegating execution to the robot \cite{pichler2017towards}. 
However, several tasks in which robot assistance would help are intrinsically human and designed for human-centric environments, and training robots to infer human goals and generate actionable plans requires extensive data collection, which remains a significant challenge \cite{brohan2023rt, o2024open, kim2024openvla}.
An alternative approach that has recently gained popularity is the use of MM-LLMs for long-term action anticipation (LTA) from observations \cite{sener2020temporal, abu2018will, abu2021long, gammulle2019forecasting}. 
Prior work has explored the in-context performance of MM-LLMs to infer a human's goal and predict the future actions needed to achieve it, based on the sequence of actions performed by the human \cite{zhao2024antgpt}.
However, directly querying an MM-LLM for planning robot actions has several limitations. 

Since MM-LLMs lack real-world interaction during pre-training, they struggle to track the effects of their planned actions \cite{bender2020climbing, ahn2022can, hao2023reasoning}.
As a result, an agent that executes the generated actions often deviates from the intended goal state.
To address this, recent work has proposed explicitly querying MM-LLMs to track the state of the system, allowing them to measure progress toward the goal \cite{liu2023reason, chen2024llm, xie2024making, yoneda2024statler, zhou2024wall}. 
However, these approaches either rely on assumptions such as the ability to comprehensively describe system states in natural language or affordably query the environment for state information; or they constrain their focus by treating state tracking separately from task planning. 
In this paper, we extend prior research on LTA by relaxing some of the assumptions and constraints to better match a real-world industrial setting.

Another key challenge in applying MM-LLMs to robotics is handling the high dimensionality of the action space. 
Since MM-LLMs fail to consider the nuances of real-world execution, several previous works limit their output to a cursory action plan of ``verb, noun'' pairs (e.g., ``pick, cup'', ``pour from, cup'', ``place on, table''), while relying on predefined functions to abstract away the other parameters \cite{ahn2022can, chen2024llm, gramopadhye2023generating, huang2022language, singh2023progprompt}. 
However, this often leads to ambiguity during action execution. 
For instance, executing ``place on, table'' with a cup requires determining a placement location on the table, while accounting for other objects, and the orientation of the cup after placement (upright, upside down, etc.).
Some works mitigate this by incorporating feedback, either from humans in the loop or by querying the environment, and re-planning for any failures. 

In this paper, we propose \system to explicitly estimate and propagate a structured representation of the environment state. 
Specifically, we consider an ubiquitous industrial scenario in which an operator is assembling a structure on a workstation and wishes to hand over the planning for building the structure to a robot. Our mock task serves as a proxy for real-world industrial tasks, such as machine assembly or modular fixture construction, where human-robot collaboration would be beneficial.
As shown in Fig. \ref{fig:teaser}, we use the actions already performed by the operator and an image of the workstation, along with a small curated set of in-context examples, to prompt an MM-LLM to estimate - the overall structure the operator is trying to build (\textit{task estimation}) (See \S \ref{sec:method_task}) and a structured state representation of the workstation in a JSON format (\textit{state representation generation}) (See \S \ref{sec:method_state}), followed by the remaining actions required to construct the structure (\textit{action prediction}) (See \S \ref{sec:method_action}) and the future states of the environment after executing the proposed actions (\textit{state propagation}) (See \S \ref{sec:method_assistance}), without requiring any interaction with the environment.
Propagating a structured representation of the environment allows us to predict assistance parameters required to complete the predicted actions that could not be expressed as a ``verb, noun'' pair.
It also allows us to track the distance from the inferred goal quantitatively. 
We use this distance metric to truncate and regenerate the future action plan and states iteratively, to ensure progress towards the inferred goal (\textit{rollouts}) (See \S \ref{sec:method_rollout}).
\system employs an MM-LLM for explicitly estimating and predicting a structured state representation to support robotic action planning in an industrial setting. Such industrial settings (1) offer a valuable opportunity for human–robot collaboration, (2) are inherently structured, making our choice of a structured state representation well-suited, and (3) remain underrepresented in existing MM-LLM pre-training datasets, posing unique challenges for model generalization.

We evaluated our approach on a dataset from a simulated block assembly scenario, in which human operators teleoperate two Franka robot arms situated on either side of a table (See \S \ref{sec:eval_exp}) \cite{prakash2025}. The goal of the assembly scenario is to construct a structure using five identical wooden blocks. 
We use several metrics (See \S \ref{sec:eval_metrics}), including executability, longest common subsequence (LCS), and final-state error to test our generated action plans.
Overall, our method increased action executability by $32.8\%$ and decreased final-state error by $14.8\%$, compared to a state-of-the-art baseline (See \S \ref{sec:results}).

\begin{figure*}[!h]
  \centering
  \includegraphics[width=0.9\linewidth]{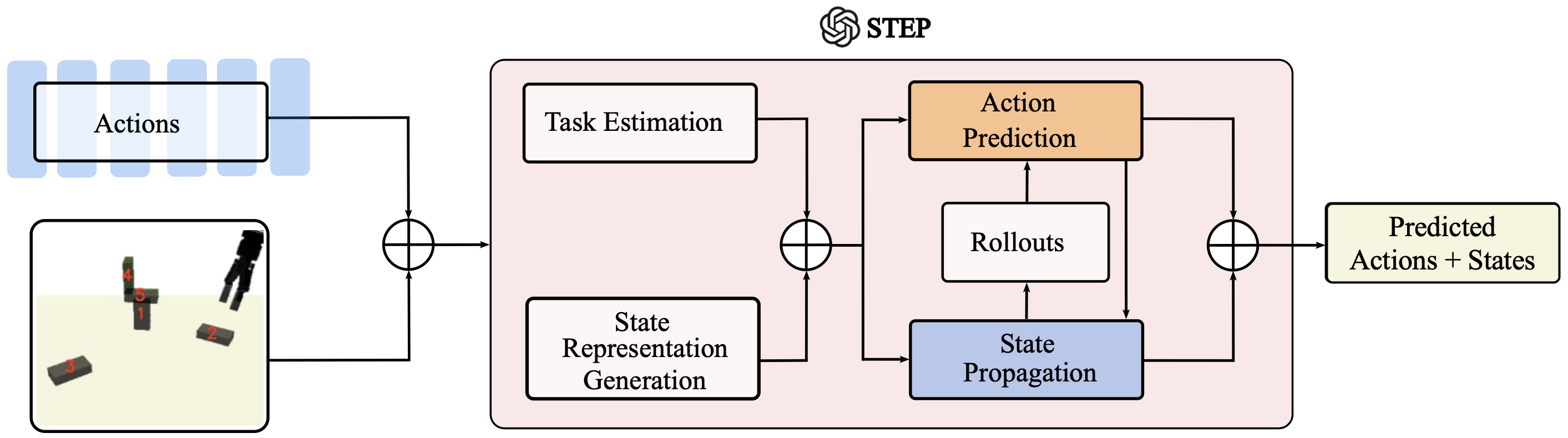}
  \caption{After a hand-over by the operator, the overall task and current system state are estimated. They are then used to iteratively rollout next actions and states multiple times. Each rollout selects the closest state to the goal to guide actions towards the estimated task.}
  \label{fig:sys_diag}
  \vspace{-0.2cm}
\end{figure*}

\section{Related Work}
\label{sec:related}
\system is inspired by papers on shared autonomy and LTA using LLMs. We address some of their limitations by extending prior work on tracking a state representation of the environment for LTA.

\subsection{Shared autonomy in teleoperation}
\label{sec:related_teleoperation}
The idea of shared teleoperation has gained popularity in industry as a way to increase production while empowering humans to assume a supervisory role \cite{pichler2017towards}.
Previous work has approached shared teleoperation by having a semi-autonomous robot provide assistance to an operator \cite{li2023classification, selvaggio2021autonomy, cai2024hierarchical}. 
Usually, the robot starts as non-autonomous and attempts to infer the user's goal \cite{yu2005telemanipulation, gao2014contextual, li2003recognition, aarno2008motion}, or form a reasonable estimate of it \cite{javdani2018shared}, by observing their actions. 
The robot then provides assistance to complete the task by taking control \cite{lin2020shared} or providing visual, haptic, or multi-modal cues \cite{manschitz2022shared, zhang2021haptic, wang2024lami, cai2024hierarchical}. 
In this paper, we follow a similar pipeline, by taking as input the actions performed by the user as a sequence of natural language actions (e.g. ``Pick block 1'', ``Stand block 1'', ``Withdraw'') and predicting the user's goal, followed by the actions required to complete the goal.

\subsection{Large language models and long-term action anticipation}
\label{sec:related_llm}
Large Language Models (LLMs), and their images + text counterparts, MM-LLMs, have shown great performance on diverse tasks \cite{davison-etal-2019-commonsense, LPAQA, Petroni2019LanguageMA, Ilharco2020ProbingTM, Cobbe2021TrainingVT, shen2021generate, lu2021fpt}. 
Particularly useful is their in-context learning ability, where LLMs used as frozen models can generalize across domains when few domain-specific examples are provided during inference \cite{brown2020language, tsimpoukelli2021multimodal}.
Several studies have proposed using in-context learning to ground LLM output in robot-actionable steps \cite{ahn2022can, gramopadhye2023generating, huang2022language, singh2023progprompt}.
In-context learning is especially useful in robot task planning for industry, as LLMs are often not trained on data that is relevant for these scenarios.
Like shared autonomy planning, LTA involves forecasting an agent’s future actions from an initial sequence of observed behavior \cite{sener2020temporal, abu2018will, abu2021long, gammulle2019forecasting}.
Previous research has explored converting a video into a sequence of natural language actions and using that as input to LLMs to predict future actions. 
AntGPT \cite{zhao2024antgpt} proposes a method in which an LLM predicts the overall goal of an agent from a sequence of natural language actions and then generates the remaining actions required to achieve that goal.
VidAssist \cite{islam2024propose} proposes to take as input the initial sequence of actions and the overall goal to iteratively build a tree of possible future action sequences and find the sequence that will best satisfy the given goal.
Inspired by these approaches, our prediction pipeline further extends this framework by also predicting and propagating the state of the environment.

\subsection{State estimation using language models}
\label{sec:related_state_est}
LLMs often struggle with planning and reasoning tasks.
Being trained for auto-regressive sequence-to-sequence translation, LLMs lack any motivation to keep track of the state of a system \cite{hao2023reasoning}. 
As a result, LLMs often hallucinate actions that diverge from the intended goal. 
To overcome this limitation, previous research has explored using chain-of-thought reasoning to explicitly track the state of the system with LLMs \cite{wei2022chain, mu2023embodiedgpt}.
RAFA \cite{liu2023reason} proposes iteratively building a tree of action and state sequences, followed by utilizing these sequences to query an LLM to select the best action to perform. 
However, RAFA requires access to the initial ground-truth state of the system to begin planning. 
Similar to RAFA, LLM-State \cite{chen2024llm} and Kaige et al., \cite{xie2024making} also output the future state representations in natural language to guide an agent's actions.
However, a natural language state representation is often not expressive enough to represent the state of a robot workspace and only allows prior works to get a vague guidance towards the goal state. 
In our method, we query an MM-LLM to estimate the environment state and propagate it as a structured JSON object, allowing us to calculate a quantitative distance from the goal state.
Statler \cite{yoneda2024statler} proposes querying an LLM to infer and update the state of a system as a structured representation. However, Statler does not consider the problem of action planning and only propagates the state for one action.
Wall-E \cite{zhou2024wall} proposes to predict future actions and propagate a state representation. However, it needs the initial ground-truth state of the system as input and an exploratory phase to learn the rules about the world, which might not be feasible in the real world.

\section{Approach}
\label{sec:method}

\system takes as input the actions performed by an operator and an image of the workstation, along with a set of example interactions to prompt an MM-LLM. As illustrated in Fig. \ref{fig:sys_diag}, our pipeline predicts the overall task the operator is trying to accomplish. Simultaneously, it also estimates a structured state representation of the current workspace from its image. It then uses the estimated task and action history to predict future actions required to complete the task. The current state is then propagated to estimate the future states of the workstation. Using the structured future state representations, it calculates their distance from the goal state of the estimated task and truncates the future actions at the state with the least distance. \system also repeats the action prediction and state propagation steps for multiple rollouts to iteratively get closer to completing the task. 

\subsection{Task Estimation}
\label{sec:method_task}

Following prior work \cite{zhao2024antgpt}, we employ a top-down approach by first estimating the overall task ($T$) that the operator wants to complete. \system takes as input the actions performed by the operator so far ($A_1, A_2, ..., A_k$) and the image of the current workstation ($I$) to prompt an MM-LLM for the task. In the prompt, we also include examples (of the format \texttt{(actions, <image>) -> Task}) from a small pre-collected dataset, and a list of all possible tasks (see \S \ref{sec:eval_exp}) to guide and constrain the MM-LLM output.

\subsection{State Representation Generation}
\label{sec:method_state}

\begin{figure}[ht]
  \centering
  \includegraphics[width=\linewidth]{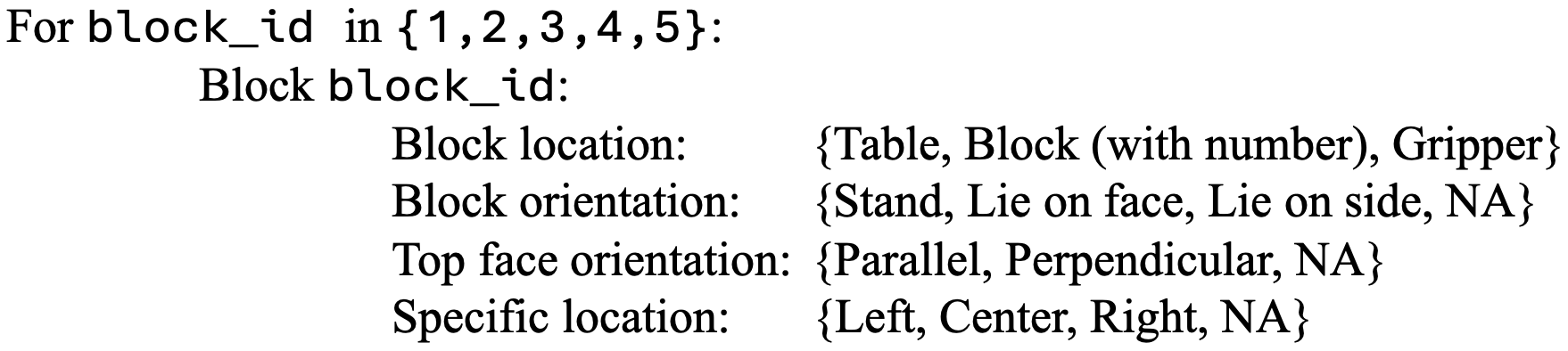}
  \caption{For each of the blocks our structured state representation records the block location (table/another block with its number), block orientation (stand/lie on face/lie on side/NA), top face orientation (long-side parallel/perpendicular to camera/NA), specific location (left/right/center if the block is on another block, otherwise NA). The modular state design also allows extension to other scenarios by adding descriptors - `nearby objects', `facing', `containing objects' etc.}
  \label{fig:state_representation}
  \vspace{-0.2cm}
\end{figure}

We prompt an MM-LLM with the image of the workstation ($I$) for a structured representation of the state ($S$). We follow a chain-of-thought approach to query the model multiple times, each time asking it to describe various parts of the workspace layout. First, we query it to get the locations (top right, top left, center right, etc.) of each of the 5 blocks in the image. These locations of the blocks in the image help us ground our subsequent prompts in the image, where we query the model to provide the following information for each block: (i) the block's placement location (including the number of the block it is placed on, if any) (ii) the block's orientation (i.e., standing, lying face down, or lying on side) (iii) the block's top face orientation (facing parallel or perpendicular to the camera) (iv) If the block is on another block, describe its location relative to the base block. We also include a brief description of the workstation and example images of the workstation in the prompts. In the end, we combine the collected information and ask an MM-LLM to format it as a JSON object (See Fig. \ref{fig:state_representation} for more details).

\subsection{Action Prediction}
\label{sec:method_action}

We use the estimated task ($T$) and the action history ($A_1, A_2, ..., A_k$) to prompt the MM-LLM for the future actions ($A_{k+1}, A_{k+2}, ..., A_n$) that need to be performed to complete the task. We also include examples (of the format - \texttt{performed actions -> future actions}) in the prompt, taken from the pre-collected dataset for completing the estimated task. During our experiments, we discovered that including multi-modal data in the prompt for this stage degraded the performance, so we did not use the state or image information for action prediction.

\subsection{State Propagation}
\label{sec:method_assistance}

We propagate the estimated current state of the workstation ($S$) to track the changes resulting from executing the predicted actions. We prompt the MM-LLM using $A_1, A_2, ..., A_k$, $S$ and $A_{k+1}$, to get as output the next state ($S_{k+1}$), also as a JSON object. We then iteratively set the propagated state as the current state and predict the states resulting from executing all predicted actions ($S_{k+1}, S_{k+2}, ..., S_n$). For example, getting $S_{k+2}$ would involve - $A_1, A_2, ..., A_k, A_{k+1}$, $S_{k+1}$, $A_{k+2}$ \texttt{->} $S_{k+2}$. Examples of the same format relevant to the estimated task are also included in the prompt. During our experiments, we found that using a chain-of-thought approach to break down the state propagation into two parts improved performance. For propagating the state at each action, we first ask the MM-LLM to describe the changes in the state in natural language, and then use the description to get a JSON representation of the next state.

\subsection{Rollouts}
\label{sec:method_rollout}

Using a structured state representation of the workstation allows us to manually calculate its distance ($d_T$) from the estimated task completion state. We calculate the distance as the total number of differences between the state and the task completion state. Since our state representation has 5 blocks and each block has 4 different characteristics, $d_T$ can range from $0$ to $20$ (See Fig. \ref{fig:state_representation}). For every rollout, we query the MM-LLM for the future actions and states. We then calculate $d_T$ for all states (current and predicted), and truncate the predicted actions and states after the state with the first instance of the lowest $d_T$ (e.g. $S_l$ could be the state with the lowest $d_T$). For the next rollout, we set $S_l$ as the current state and use $A_1, ..., A_k, ..., A_l$ and $S_l$ for action prediction and state propagation. Since $S_l$ has a $d_T$ equal to or lower than $S$, 
with each rollout \system identifies a state closer to the inferred goal than the state at the start of the rollout, and helps guide the action plan towards the goal. 
We repeat the rollouts for $max\_rollout$ times, which is a hyperparameter, or until we reach a state with $d_T = 0$. After the final rollout, we return the predicted action sequence ($A_{k+1}, ..., A_l$) and state sequence ($S, S_{k+1}, ..., S_l$), where $d_T$ of $S_l$ is the lowest among all states.


\section{Evaluation}
\label{sec:evaluation}
We compared our method with a state-of-the-art LTA baseline inspired by the top-down approach proposed by AntGPT \cite{zhao2024antgpt}. The baseline consisted of using the same inputs as our method, for \textit{task estimation} (\S \ref{sec:method_task}) followed by a single rollout of \textit{action prediction} (\S \ref{sec:method_action}), with no state information. 
We selected this comparison to directly examine the impact of explicitly estimating and tracking the system state on action prediction accuracy. To isolate and study this effect, we deliberately excluded potential confounding factors such as interactive planning or human-in-the-loop feedback. 
Also, several related works we examined impose constraints such as requiring access to the initial ground-truth state of the system for planning, an assumption that is often unrealistic in real-world industrial human–robot collaboration scenarios. Moreover, many approaches address only specific subcomponents of the overall action planning problem rather than the complete pipeline. Considering these factors, we identified AntGPT \cite{zhao2024antgpt} as a suitable inspiration for our baseline.


\subsection{Experimental Setup and Dataset}
\label{sec:eval_exp}

\begin{figure}[ht]
  \centering
  \includegraphics[width=\linewidth]{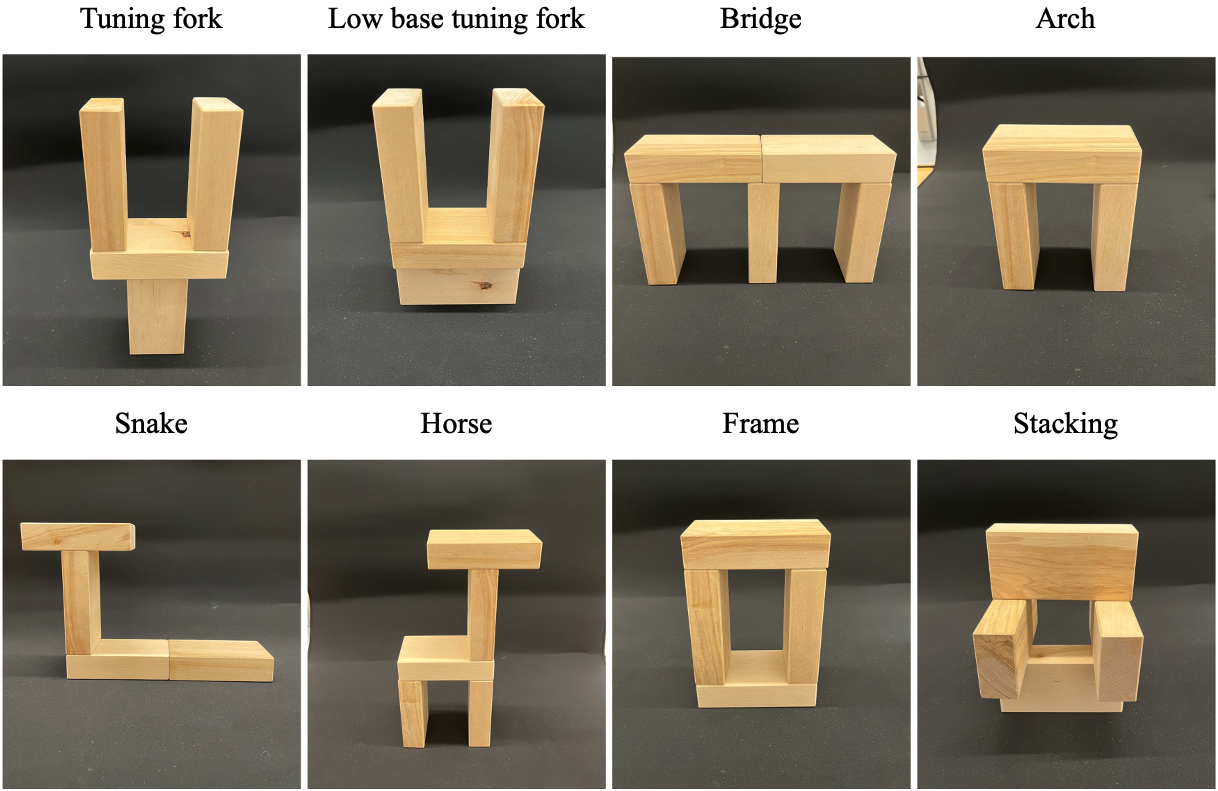}
  \caption{The eight block assembly tasks.}
  \label{fig:goals}
  \vspace{-0.2cm}
\end{figure}

We evaluated our method on a dataset of teleoperated manipulation sequences collected by Baskaran et al. \cite{prakash2025}. The collected manipulation sequences involved users teleoperating two robots in real time within a simulated environment using HTC Vive Pro controllers \cite{htc_vive_pro_controllers}. The simulated workstation composed of a table mounted with two Franka Emika robot arms \cite{franka} on either side, and five identical wooden block assembly pieces.
The collected dataset had 495 instances from 19 users, where each instance involved operators accomplishing the goal of assembling the blocks into one of the $8$ different structures shown in Fig. \ref{fig:goals} - \texttt{Tuning fork, Low base tuning fork, Bridge, Arch, Snake, Horse, Frame and Stacking}. Please see Baskaran et al. \cite{prakash2025} for more details on the dataset collection procedure. We manually annotated the dataset instances to add labels and post-processed them to get the following features per instance: Task ($T$), Recorded actions ($A_1, A_2, ..., A_n$), Images of the workspace ($I_0, I_1, ..., I_n$), States of the workspace ($S_0, S_1, ..., S_n$) and Possible tasks ($Tp_0, Tp_1, ..., Tp_n$). $T$ denotes the goal structure. A recorded action $A_i$ is of the format \texttt{<Left robot action ($A_{il}$), Right robot action ($A_{ir}$)>}. $A_{il}$ and $A_{ir}$ can be one of the nine actions - \texttt{Pick block $B_x$}, \texttt{Stand block $B_x$}, \texttt{Lie block $B_x$}, \texttt{Side-lie block $B_x$}, \texttt{Stand-on-block block $B_x$}, \texttt{Lie-on-block block $B_x$}, \texttt{Side-lie-on-block block $B_x$}, \texttt{Idle}, and \texttt{Withdraw}. An image $I_i$ is taken from the video where $A_i$ ends and $A_{i+1}$ begins. We also added block numbers to the images in post-processing to distinguish the blocks (See Fig. \ref{fig:sys_diag}). $S_i$ is the structured state representation of $I_i$. $Tp_i$ refers to all possible structures that share $S_i$ as an intermediate state, as some tasks may overlap with others. For example, the Arch task overlaps with the Bridge and Horse tasks (see Fig. \ref{fig:goals}). 
For our evaluation we used the initial part of the recorded actions ($A_1, A_2, ..., A_k$) and the image of the workstation after completing those actions ($I_k$), to predict the remaining actions required to complete the task ($T$).


\subsection{Metrics}
\label{sec:eval_metrics}

To calculate quantitative metrics for the predicted actions, we first manually propagate the ground-truth current state of the workstation using hand-crafted rules. Our rules take as input the predicted actions, and the predicted states which help to determine assistance parameters to remove ambiguity from action execution. The baseline (See \S \ref{sec:evaluation}) does not predict future states and thus suffers from ambiguity in execution. So we randomly select a state from the set of possible states that can result from executing the actions predicted by the baseline. Note that here we select a state only from those reachable by the predicted actions, not from all possible states. Following existing research \cite{gramopadhye2023generating, huang2022language}, we show results across three metrics: executability, final-state error, and longest common subsequence. We also provide the task correctness and current state error to measure the correctness in predicting the overall task and the error in inferring the current state of the workspace.

\textbf{Executability} ensures that the predicted actions follow a logical order and satisfy execution constraints (e.g., proper object poses, action preconditions, etc.). We perform checks for executability and truncate the predicted action sequence at the first action that fails the checks. We report executability as the percentage of predicted actions that passed all the checks.

\textbf{Final-state error} measures the error of the ground truth task completion state from the final state that could be successfully achieved by executing the predicted actions. For the final achievable state, we select the resulting state from the last predicted action that can be executed successfully. This metric measures the difference between the result and the expected goal.

\textbf{Longest Common Sub-sequence (LCS)} is defined as the length of the longest common sub-sequence of actions in the predicted actions and the remaining recorded actions from our dataset, divided by the length of the longer sequence between predicted actions and remaining recorded actions. Following prior work \cite{gramopadhye2023generating, huang2022language, puig2018virtualhome}, we allowed gaps to exist between the actions in the common sub-sequence, provided their order was the same. We report LCS as a percentage. A high LCS can mean that the predicted actions are relevant to the task and have short-term order. However, it does not provide a complete picture of correctness on its own, as there can be multiple correct action sequences with varying LCS values. A high LCS also does not guarantee that the predicted actions will execute successfully.

\textbf{Task correctness} measures the accuracy of \textit{task estimation} (See \S \ref{sec:method_task}). It is assigned $1$ if the predicted task matches any of the possible tasks at inference time; otherwise, it is set to $0$.

\textbf{Current state error} compares the distance between the current state output of \textit{state representation generation} (See \S \ref{sec:method_state}) and the ground-truth state at the time of inference. 



\section{Results \& Discussion}
\label{sec:results}
\begin{table*}[!h]
    \vspace{0.2cm}
    \caption{Comparison of executability, final error, and LCS of \system with the baseline, along with task correctness and current state error for multiple task completion percentages.}
    \begin{center}
    \includegraphics[width=0.8\linewidth]{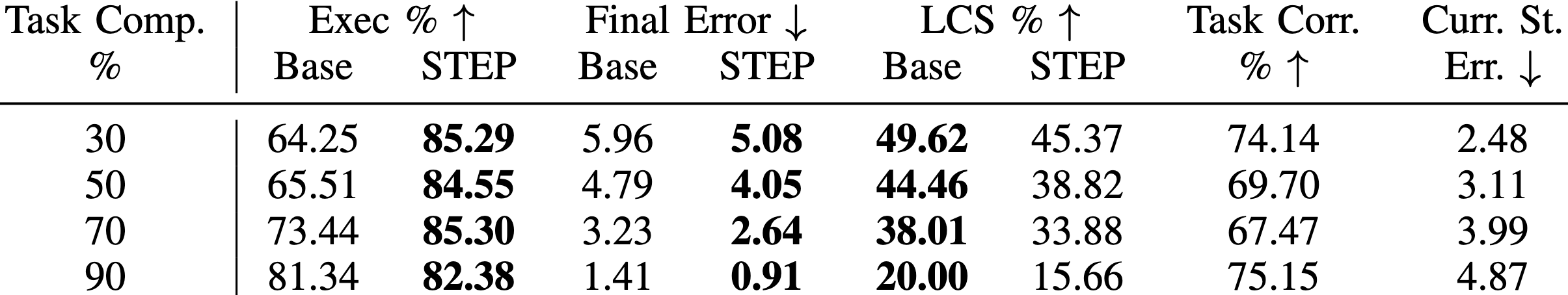}
    \label{fig:result_main}
    \end{center}
    \vspace{-0.2cm}
\end{table*}


We used the GPT-4o \cite{chatGPT} model
offered by the OpenAI API \cite{OpenAI_API} for our main results. Every time we prompt the MM-LLM, we sample $5$ outputs and pick the output with the maximum log-probability returned by the model.

\subsection{Comparison with Baseline}
\label{sec:main_res}

To model variations in user delegations to \system, we conduct experiments at multiple task completion percentages of 30\%, 50\%, 70\%, and 90\%. A task completion of $x$\% signifies that the initial $x$\% of the actions from the recorded action sequence were used as input for evaluation. 


As can be seen in Table \ref{fig:result_main}, \system predicted action plans that were more executable than the baseline, for all percentages of task completion. Also, by predicting future states along with actions, our method resulted in a state that was closer to the intended goal of the operator, when compared to the baseline. However, we noted that the baseline method had a higher LCS than our method. This behavior can be attributed to STEP truncating the predicted actions after the state with the least distance from the goal. As a result, our predicted action sequence was on average approximately $3$ actions shorter in length than the baseline's predictions. We suspect that the extra actions produced by the baseline caused it to have more number of actions in common with the recorded actions, thus increasing the baseline LCS. However, the higher LCS of the baseline resulted in unnecessary actions that were less executable and less correct in the resultant state. Action plans generated by \system were more efficient, as they had a lower final error compared to the baseline, while being shorter on average. We hypothesize that noise in the recorded actions, caused by operator errors when collecting the data also contributed to LCS being a questionable metric in our evaluation.

We noticed that with increasing task completion percentage, both the methods achieved a better final error, since the robot started from a state closer to the goal state, as more of the assembly was completed by the operator. With increasing task completion, the task correctness first declines due to increasing noise in longer input action sequences and more occlusion of the block structures in the images, and then increases as the structure being built becomes more discernible. We suspect that the current state error increased as the task progressed because the structure became more complicated and blocks got closer together in the image.

\subsection{Ablations}
\label{sec:ablations}

\begin{table}[h!]
\centering
\caption{Ablation of rollouts for 50\% task completion. Final error improves with rollouts showing that additional rollouts decrease the distance from goal.}
\label{fig:ablate_rollout}
\normalsize
\begin{tabular}{ c | c c c}
 Rollouts & Exec. \% $\uparrow$ & Final Err. $\downarrow$ & LCS \% $\uparrow$\\
\midrule
1 & \textbf{88.98} & 4.40 & 31.06\\
2 & 85.60 & 4.11 & 37.4\\
3 & 84.55 & \textbf{4.05} & \textbf{38.82}\\
\end{tabular}
\vspace{-0.1cm}
\end{table}

We ablated the number of rollouts conducted before calculating the metrics. Table \ref{fig:ablate_rollout} shows the effect of the number of rollouts on performance for 50\% task completion. We noticed that each rollout predicted a longer action sequence than the preceding rollout, with action plan length increasing by $0.7$ actions on average with each rollout. LCS increased with more rollouts, which is explained by the increased number of actions in the action plan. The increasing actions did cause the executability to decrease. However, as seen from the decreasing final error with more rollouts, the added actions that could execute were necessary as they were able to result in a final state closer to the goal with increasing rollouts.

\begin{table}[h!]
\centering
\caption{Ablating MM-LLM to compare executability and final error of \system with the baseline for 50\% task completion. Results for $45$ random dataset instances.}
\label{fig:ablate_mmllm}
\begin{tabular}{ c | c c c c c}
 Model & \multicolumn{2}{c}{Exec. \% $\uparrow$} & \multicolumn{2}{c}{Final Error $\downarrow$} & Task Corr.\\ 
 & Base & \system & Base & \system & \% $\uparrow$\\
\midrule
GPT-4o & 71.68 & \textbf{78.12} & 4.53 & \textbf{4.38} & 62.22\\
GPT-4o mini & \textbf{82.60} & 82.16 & \textbf{4.49} & 4.93 & 26.67\\
GPT-4.1 & 82.28 & \textbf{92.02} & 4.80 & \textbf{4.11} & 71.11\\
GPT-5.1 & 69.24 & \textbf{82.70} & 4.80 & \textbf{3.71} & 73.33\\
GPT-5.2 & 76.64 & \textbf{84.44} & 4.73 & \textbf{4.49} & 71.11\\
\end{tabular}
\end{table}

We compared \system with the baseline for various MM-LLMs - GPT-4o, GPT-4o mini, GPT-4.1, GPT-5.1, and GPT-5.2, offered by the OpenAI API \cite{OpenAI_API}. These models were selected to cover a range of model sizes and generation paradigms (regular LLMs and reasoning LLMs). To reduce computational cost and runtime, this ablation was conducted on a representative subset of $45$ dataset instances. These instances were randomly sampled from our dataset while ensuring coverage across all eight task categories (see Fig. \ref{fig:goals}). As seen in Table \ref{fig:ablate_mmllm}, \system outperformed the baseline across most models in terms of both executability and final error, demonstrating its robustness and generalizability across different MM-LLMs. For GPT‑4o mini, however, the baseline achieved slightly better results than \system. We observed that GPT‑4o mini exhibited a notably low task correctness score, which measures the task estimation stage, the first step in our pipeline, and consequently propagated errors throughout subsequent stages for both methods. As a result, the outputs corresponded to an incorrect task, rendering these results less reliable compared to those from other models. We attribute this behavior to GPT‑4o mini’s relatively smaller model size, which likely limits its task estimation capabilities.

\begin{table}[h!]
\centering
\caption{Ablation of task correctness for 50\% task completion. All metrics improve with correct task prediction. Improvement in earlier stages improves downstream performance.}
\label{fig:ablate_task_gt}
\normalsize
\begin{tabular}{ c | c c c}
 Task corr. & Exec. \% $\uparrow$ & Final Err. $\downarrow$ & LCS \% $\uparrow$\\
\midrule
1 & \textbf{84.63} & \textbf{3.64} & \textbf{41.16}\\
0 & 84.37 & 4.9 & 33.45\\
\end{tabular}
\vspace{-0.1cm}
\end{table}

\begin{table}[h!]
\centering
\caption{Ablation of current state error for 50\% task completion. Executability and final error improve with no error in the current state estimation.}
\label{fig:ablate_scene_gt}
\normalsize
\begin{tabular}{ c | c c c}
 Curr. st. err. & Exec. \% $\uparrow$ & Final Err. $\downarrow$ & LCS \% $\uparrow$\\
\midrule
0 & \textbf{86.38} & \textbf{3.93} & 36.08\\
$>$0 & 84.41 & 4.06 & \textbf{39.07}\\
\end{tabular}
\vspace{-0.1cm}
\end{table}

We also studied the impact of task and state estimation on action prediction and state propagation. Table \ref{fig:ablate_task_gt} compares the results for the data instances that had a correct task prediction with the instances with an incorrect task prediction. Similarly, Table \ref{fig:ablate_scene_gt} shows the results for instances that did not have any errors in their current state prediction compared to the other instances. On average, we saw an improvement in performance of the predicted actions if the task and current state were predicted correctly. This ablation also shows the modularity of our approach as stages can be worked on independently and an improved performance earlier in the pipeline can reliably be transferred to better results for the final predicted actions.

\section{Conclusion}
\label{sec:conclusion}
We introduce \system, a method to enhance long-term action anticipation with MM-LLMs by explicitly modeling state transitions, leading to more accurate and executable robot action plans in collaborative industrial tasks. We construct a multi-staged pipeline composed of first estimating the overall task a robot operator is trying to perform, along with a structured representation of the current state of the robot's workstation. We then pass this information on to later stages to predict the actions required to complete the task and propagate the state representation to obtain the assistance parameters required to successfully execute the predicted actions for the desired task. We also track the distance between the propagated states and the goal state and guide predicted actions towards the goal. 

We extensively evaluated our method on a dataset of an assembly scenario with teleoperated robots. Our results highlight our contribution towards bridging some of the gaps in the existing literature by demonstrating improved performance in executability and final error over the existing state-of-the-art. However, we see some areas for improvement. One limitation of \system is that its current implementation is specific to the block assembly scenario. Industry may extend \system to other scenarios by using information relevant to the new scenarios, by modifying the prompts, state representation structure, and function implementations. This investment in modifying \system for a new scenario would only be required once at the start. However, future work can explore ways to generalize \system to various scenarios with minimal modifications. A potential avenue of research could be using an MM-LLM to analyze a scenario and automatically craft the required prompts, state representation structure, and function implementations.

Additional work is also required to make \system real-time before it can be adopted for practical use. In order to propagate states and guide the generation of action plans towards the goal, \system queries an MM-LLM several more times than the baseline. However, this causes \system to trade off speed for performance accuracy. In our experiments, we observed that the baseline had a shorter run time (M = 18.24, SD = 4.23 sec) than \system (M = 148.07, SD = 55.03 sec). Future research could investigate hosting an MM-LLM locally or finetuning a smaller model to reduce the latency of each query. We hope that our work encourages further inquiries into the potential of MM-LLMs in long-horizon reasoning for human-robot collaboration.




\bibliographystyle{IEEEtran}
\bibliography{example}  

@article{pichler2017towards,
  title={Towards shared autonomy for robotic tasks in manufacturing},
  author={Pichler, Andreas and Akkaladevi, Sharath Chandra and Ikeda, Markus and Hofmann, Michael and Plasch, Matthias and W{\"o}gerer, Christian and Fritz, Gerald},
  journal={Procedia Manufacturing},
  volume={11},
  pages={72--82},
  year={2017},
  publisher={Elsevier}
}

@article{li2023classification,
  title={The classification and new trends of shared control strategies in telerobotic systems: A survey},
  author={Li, Gaofeng and Li, Qiang and Yang, Chenguang and Su, Yuan and Yuan, Zuqiang and Wu, Xinyu},
  journal={IEEE Transactions on Haptics},
  volume={16},
  number={2},
  pages={118--133},
  year={2023},
  publisher={IEEE}
}

@article{selvaggio2021autonomy,
  title={Autonomy in physical human-robot interaction: A brief survey},
  author={Selvaggio, Mario and Cognetti, Marco and Nikolaidis, Stefanos and Ivaldi, Serena and Siciliano, Bruno},
  journal={IEEE Robotics and Automation Letters},
  volume={6},
  number={4},
  pages={7989--7996},
  year={2021},
  publisher={IEEE}
}

@inproceedings{yu2005telemanipulation,
  title={Telemanipulation assistance based on motion intention recognition},
  author={Yu, Wentao and Alqasemi, Redwan and Dubey, Rajiv and Pernalete, Norali},
  booktitle={Proceedings of the 2005 IEEE international conference on robotics and automation},
  pages={1121--1126},
  year={2005},
  organization={IEEE}
}

@article{javdani2018shared,
  title={Shared autonomy via hindsight optimization for teleoperation and teaming},
  author={Javdani, Shervin and Admoni, Henny and Pellegrinelli, Stefania and Srinivasa, Siddhartha S and Bagnell, J Andrew},
  journal={The International Journal of Robotics Research},
  volume={37},
  number={7},
  pages={717--742},
  year={2018},
  publisher={SAGE Publications Sage UK: London, England}
}

@inproceedings{gao2014contextual,
  title={Contextual task-aware shared autonomy for assistive mobile robot teleoperation},
  author={Gao, Ming and Oberl{\"a}nder, Jan and Schamm, Thomas and Z{\"o}llner, J Marius},
  booktitle={2014 IEEE/RSJ International Conference on Intelligent Robots and Systems},
  pages={3311--3318},
  year={2014},
  organization={IEEE}
}

@inproceedings{li2003recognition,
  title={Recognition of operator motions for real-time assistance using virtual fixtures},
  author={Li, Ming and Okamura, Allison M},
  booktitle={11th Symposium on Haptic Interfaces for Virtual Environment and Teleoperator Systems, 2003. HAPTICS 2003. Proceedings.},
  pages={125--131},
  year={2003},
  organization={IEEE}
}

@article{aarno2008motion,
  title={Motion intention recognition in robot assisted applications},
  author={Aarno, Daniel and Kragic, Danica},
  journal={Robotics and Autonomous Systems},
  volume={56},
  number={8},
  pages={692--705},
  year={2008},
  publisher={Elsevier}
}

@inproceedings{lin2020shared,
  title={Shared autonomous interface for reducing physical effort in robot teleoperation via human motion mapping},
  author={Lin, Tsung-Chi and Krishnan, Achyuthan Unni and Li, Zhi},
  booktitle={2020 IEEE International Conference on Robotics and Automation (ICRA)},
  pages={9157--9163},
  year={2020},
  organization={IEEE}
}

@inproceedings{manschitz2022shared,
  title={Shared Autonomy for Intuitive Teleoperation},
  author={Manschitz, Simon and Ruiken, Dirk},
  booktitle={ICRA Workshop: Shared Autonomy in Physical Human-Robot Interaction: Adaptability and Trust},
  year={2022}
}

@inproceedings{zhang2021haptic,
  title={Haptic feedback improves human-robot agreement and user satisfaction in shared-autonomy teleoperation},
  author={Zhang, Dawei and Tron, Roberto and Khurshid, Rebecca P},
  booktitle={2021 ieee international conference on robotics and automation (icra)},
  pages={3306--3312},
  year={2021},
  organization={IEEE}
}

@inproceedings{wang2024lami,
  title={LaMI: Large Language Models for Multi-Modal Human-Robot Interaction},
  author={Wang, Chao and Hasler, Stephan and Tanneberg, Daniel and Ocker, Felix and Joublin, Frank and Ceravola, Antonello and Deigmoeller, Joerg and Gienger, Michael},
  booktitle={Extended Abstracts of the CHI Conference on Human Factors in Computing Systems},
  pages={1--10},
  year={2024}
}

@article{cai2024hierarchical,
  title={Hierarchical Deep Learning for Intention Estimation of Teleoperation Manipulation in Assembly Tasks},
  author={Cai, Mingyu and Patel, Karankumar and Iba, Soshi and Li, Songpo},
  journal={arXiv preprint arXiv:2403.19770},
  year={2024}
}

@inproceedings{davison-etal-2019-commonsense,
    title = "Commonsense Knowledge Mining from Pretrained Models",
    author = "Davison, Joe  and
      Feldman, Joshua  and
      Rush, Alexander",
    booktitle = "Proceedings of the 2019 Conference on Empirical Methods in Natural Language Processing and the 9th International Joint Conference on Natural Language Processing (EMNLP-IJCNLP)",
    month = nov,
    year = "2019",
    address = "Hong Kong, China",
    publisher = "Association for Computational Linguistics",
    url = "https://aclanthology.org/D19-1109",
    doi = "10.18653/v1/D19-1109",
    pages = "1173--1178",
}

@article{LPAQA,
  author    = {Zhengbao Jiang and
               Frank F. Xu and
               Jun Araki and
               Graham Neubig},
  title     = {How Can We Know What Language Models Know?},
  journal   = {CoRR},
  volume    = {abs/1911.12543},
  year      = {2019},
  url       = {http://arxiv.org/abs/1911.12543},
  eprinttype = {arXiv},
  eprint    = {1911.12543},
  bibsource = {dblp computer science bibliography, https://dblp.org}
}

@article{Petroni2019LanguageMA,
  title={{Language Models as Knowledge Bases?}},
  author={Fabio Petroni and Tim Rockt{\"a}schel and Patrick Lewis and Anton Bakhtin and Yuxiang Wu and Alexander H. Miller and Sebastian Riedel},
  journal={ArXiv},
  year={2019},
  volume={abs/1909.01066}
}

@article{Ilharco2020ProbingTM,
  title={{Probing Text Models for Common Ground with Visual Representations}},
  author={Gabriel Ilharco and Rowan Zellers and Ali Farhadi and Hannaneh Hajishirzi},
  journal={ArXiv},
  year={2020},
  volume={abs/2005.00619}
}

@article{Cobbe2021TrainingVT,
  title={{Training Verifiers to Solve Math Word Problems}},
  author={Karl Cobbe and Vineet Kosaraju and Mohammad Bavarian and Jacob Hilton and Reiichiro Nakano and Christopher Hesse and John Schulman},
  journal={ArXiv},
  year={2021},
  volume={abs/2110.14168}
}

@article{shen2021generate,
  title={{Generate \& rank: A multi-task framework for math word problems}},
  author={Shen, Jianhao and Yin, Yichun and Li, Lin and Shang, Lifeng and Jiang, Xin and Zhang, Ming and Liu, Qun},
  journal={arXiv preprint arXiv:2109.03034},
  year={2021}
}

@article{lu2021fpt,
  title={{Pretrained Transformers as Universal Computation Engines}},
  author={Kevin Lu and Aditya Grover and Pieter Abbeel and Igor Mordatch},
  journal={arXiv preprint arXiv:2103.05247},
  year={2021}
}

@article{tsimpoukelli2021multimodal,
  title={{Multimodal few-shot learning with frozen language models}},
  author={Tsimpoukelli, Maria and Menick, Jacob and Cabi, Serkan
          and Eslami, SM and Vinyals, Oriol and Hill, Felix},
  journal={Proc. Neural Information Processing Systems},
  year={2021}
}

@article{brown2020language,
  title={Language models are few-shot learners},
  author={Brown, Tom and Mann, Benjamin and Ryder, Nick and Subbiah, Melanie and Kaplan, Jared D and Dhariwal, Prafulla and Neelakantan, Arvind and Shyam, Pranav and Sastry, Girish and Askell, Amanda and others},
  journal={Advances in neural information processing systems},
  volume={33},
  pages={1877--1901},
  year={2020}
}

@misc{gramopadhye2023generating,
      title={Generating Executable Action Plans with Environmentally-Aware Language Models}, 
      author={Maitrey Gramopadhye and Daniel Szafir},
      year={2023},
      eprint={2210.04964},
      archivePrefix={arXiv},
      primaryClass={cs.RO}
}

@inproceedings{singh2023progprompt,
  title={Progprompt: Generating situated robot task plans using large language models},
  author={Singh, Ishika and Blukis, Valts and Mousavian, Arsalan and Goyal, Ankit and Xu, Danfei and Tremblay, Jonathan and Fox, Dieter and Thomason, Jesse and Garg, Animesh},
  booktitle={2023 IEEE International Conference on Robotics and Automation (ICRA)},
  pages={11523--11530},
  year={2023},
  organization={IEEE}
}

@article{ahn2022can,
  title={Do as i can, not as i say: Grounding language in robotic affordances},
  author={Ahn, Michael and Brohan, Anthony and Brown, Noah and Chebotar, Yevgen and Cortes, Omar and David, Byron and Finn, Chelsea and Fu, Chuyuan and Gopalakrishnan, Keerthana and Hausman, Karol and others},
  journal={arXiv preprint arXiv:2204.01691},
  year={2022}
}

@inproceedings{huang2022language,
  title={Language models as zero-shot planners: Extracting actionable knowledge for embodied agents},
  author={Huang, Wenlong and Abbeel, Pieter and Pathak, Deepak and Mordatch, Igor},
  booktitle={International Conference on Machine Learning},
  pages={9118--9147},
  year={2022},
  organization={PMLR}
}

@inproceedings{sener2020temporal,
  title={Temporal aggregate representations for long-range video understanding},
  author={Sener, Fadime and Singhania, Dipika and Yao, Angela},
  booktitle={Computer Vision--ECCV 2020: 16th European Conference, Glasgow, UK, August 23--28, 2020, Proceedings, Part XVI 16},
  pages={154--171},
  year={2020},
  organization={Springer}
}

@inproceedings{abu2018will,
  title={When will you do what?-anticipating temporal occurrences of activities},
  author={Abu Farha, Yazan and Richard, Alexander and Gall, Juergen},
  booktitle={Proceedings of the IEEE conference on computer vision and pattern recognition},
  pages={5343--5352},
  year={2018}
}

@inproceedings{abu2021long,
  title={Long-term anticipation of activities with cycle consistency},
  author={Abu Farha, Yazan and Ke, Qiuhong and Schiele, Bernt and Gall, Juergen},
  booktitle={Pattern Recognition: 42nd DAGM German Conference, DAGM GCPR 2020, T{\"u}bingen, Germany, September 28--October 1, 2020, Proceedings 42},
  pages={159--173},
  year={2021},
  organization={Springer}
}

@article{gammulle2019forecasting,
  title={Forecasting future action sequences with neural memory networks},
  author={Gammulle, Harshala and Denman, Simon and Sridharan, Sridha and Fookes, Clinton},
  journal={arXiv preprint arXiv:1909.09278},
  year={2019}
}

@inproceedings{
zhao2024antgpt,
title={Ant{GPT}: Can Large Language Models Help Long-term Action Anticipation from Videos?},
author={Qi Zhao and Shijie Wang and Ce Zhang and Changcheng Fu and Minh Quan Do and Nakul Agarwal and Kwonjoon Lee and Chen Sun},
booktitle={The Twelfth International Conference on Learning Representations},
year={2024},
url={https://openreview.net/forum?id=Bb21JPnhhr}
}

@inproceedings{islam2024propose,
  title={Propose, Assess, Search: Harnessing LLMs for Goal-Oriented Planning in Instructional Videos},
  author={Islam, Md Mohaiminul and Nagarajan, Tushar and Wang, Huiyu and Chu, Fu-Jen and Kitani, Kris and Bertasius, Gedas and Yang, Xitong},
  booktitle={European Conference on Computer Vision},
  pages={436--452},
  year={2024},
  organization={Springer}
}

@article{hao2023reasoning,
  title={Reasoning with language model is planning with world model},
  author={Hao, Shibo and Gu, Yi and Ma, Haodi and Hong, Joshua Jiahua and Wang, Zhen and Wang, Daisy Zhe and Hu, Zhiting},
  journal={arXiv preprint arXiv:2305.14992},
  year={2023}
}

@article{wei2022chain,
  title={Chain-of-thought prompting elicits reasoning in large language models},
  author={Wei, Jason and Wang, Xuezhi and Schuurmans, Dale and Bosma, Maarten and Xia, Fei and Chi, Ed and Le, Quoc V and Zhou, Denny and others},
  journal={Advances in neural information processing systems},
  volume={35},
  pages={24824--24837},
  year={2022}
}

@article{mu2023embodiedgpt,
  title={Embodiedgpt: Vision-language pre-training via embodied chain of thought},
  author={Mu, Yao and Zhang, Qinglong and Hu, Mengkang and Wang, Wenhai and Ding, Mingyu and Jin, Jun and Wang, Bin and Dai, Jifeng and Qiao, Yu and Luo, Ping},
  journal={arXiv preprint arXiv:2305.15021},
  year={2023}
}

@article{liu2023reason,
      title={Reason for Future, Act for Now: A Principled Framework for Autonomous LLM Agents with Provable Sample Efficiency},
      author={Liu, Zhihan and Hu, Hao and Zhang, Shenao and Guo, Hongyi and Ke, Shuqi and Liu, Boyi and Wang, Zhaoran},
      journal={arXiv preprint arXiv:2309.17382},
      year={2023}
}

@article{chen2024llm,
  title={Llm-state: Open world state representation for long-horizon task planning with large language model},
  author={Chen, Siwei and Xiao, Anxing and Hsu, David},
  journal={arXiv preprint arXiv:2311.17406},
  year={2024}
}

@article{xie2024making,
  title={Making large language models into world models with precondition and effect knowledge},
  author={Xie, Kaige and Yang, Ian and Gunerli, John and Riedl, Mark},
  journal={arXiv preprint arXiv:2409.12278},
  year={2024}
}

@inproceedings{yoneda2024statler,
  title={Statler: State-maintaining language models for embodied reasoning},
  author={Yoneda, Takuma and Fang, Jiading and Li, Peng and Zhang, Huanyu and Jiang, Tianchong and Lin, Shengjie and Picker, Ben and Yunis, David and Mei, Hongyuan and Walter, Matthew R},
  booktitle={2024 IEEE International Conference on Robotics and Automation (ICRA)},
  pages={15083--15091},
  year={2024},
  organization={IEEE}
}

@article{zhou2024wall,
  title={Wall-e: World alignment by rule learning improves world model-based llm agents},
  author={Zhou, Siyu and Zhou, Tianyi and Yang, Yijun and Long, Guodong and Ye, Deheng and Jiang, Jing and Zhang, Chengqi},
  journal={arXiv preprint arXiv:2410.07484},
  year={2024}
}

@inproceedings{puig2018virtualhome,
  title={Virtualhome: Simulating household activities via programs},
  author={Puig, Xavier and Ra, Kevin and Boben, Marko and Li, Jiaman and Wang, Tingwu and Fidler, Sanja and Torralba, Antonio},
  booktitle={Proceedings of the IEEE conference on computer vision and pattern recognition},
  pages={8494--8502},
  year={2018}
}

@article{chatGPT, 
 author = {OpenAI},
 title = {{Hello GPT-4o}},
 url = {https://openai.com/index/hello-gpt-4o/},
 year = {2024}
}

@article{OpenAI_API, 
 author = {Brockman, Greg and Welinder, Peter and Murati, Mira and OpenAI},
 title = {OpenAI: OpenAI API},
 url = {https://openai.com/blog/openai-api},
 year = {2020}
}

@article{brohan2023rt,
  title={Rt-2: Vision-language-action models transfer web knowledge to robotic control},
  author={Brohan, Anthony and Brown, Noah and Carbajal, Justice and Chebotar, Yevgen and Chen, Xi and Choromanski, Krzysztof and Ding, Tianli and Driess, Danny and Dubey, Avinava and Finn, Chelsea and others},
  journal={arXiv preprint arXiv:2307.15818},
  year={2023}
}

@article{kim2024openvla,
  title={Openvla: An open-source vision-language-action model},
  author={Kim, Moo Jin and Pertsch, Karl and Karamcheti, Siddharth and Xiao, Ted and Balakrishna, Ashwin and Nair, Suraj and Rafailov, Rafael and Foster, Ethan and Lam, Grace and Sanketi, Pannag and others},
  journal={arXiv preprint arXiv:2406.09246},
  year={2024}
}

@inproceedings{o2024open,
  title={Open x-embodiment: Robotic learning datasets and rt-x models: Open x-embodiment collaboration 0},
  author={O’Neill, Abby and Rehman, Abdul and Maddukuri, Abhiram and Gupta, Abhishek and Padalkar, Abhishek and Lee, Abraham and Pooley, Acorn and Gupta, Agrim and Mandlekar, Ajay and Jain, Ajinkya and others},
  booktitle={2024 IEEE International Conference on Robotics and Automation (ICRA)},
  pages={6892--6903},
  year={2024},
  organization={IEEE}
}

@inproceedings{bender2020climbing,
  title={Climbing towards NLU: On meaning, form, and understanding in the age of data},
  author={Bender, Emily M and Koller, Alexander},
  booktitle={Proceedings of the 58th annual meeting of the association for computational linguistics},
  pages={5185--5198},
  year={2020}
}

@misc{franka,
  author       = {Franka Emika GmbH.},
  title        = {Franka Emika Robot: Instruction Handbook},
  year         = {2021}
}

@misc{htc_vive_pro_controllers,
  author       = {HTC Corporation},
  title        = {HTC Vive Pro Controllers},
  year         = {2025},
  howpublished = {\url{https://www.vive.com/}},
}

@inproceedings{prakash2025,
  title={e{X}plainable Intention Estimation in Teleoperated Manipulation Using Deep Dynamic Graph Neural Networks},
  author={Baskaran, Prakash and Liu, Xiao and Li, Songpo and Iba, Soshi},
  booktitle={2025 IEEE/RSJ International Conference on Intelligent Robots and Systems (IROS)},
  year={2025},
  organization={IEEE}
}
\end{document}